# DL-AMP and DBTO: An Automatic Merge Planning and Trajectory Optimization and its Application in Autonomous Driving

Yuncheng Jiang[1], Qi Lin[1], Jiwei Zhang, Jun Wang, Danjian Qian, Yuxi Cai

*Abstract*— This paper presents an automatic merging algorithm for autonomous driving vehicles, which decouples the specific motion planning problem into a Dual-Layer Automatic Merge Planning (DL-AMP) and a Descent-Based Trajectory Optimization (DBTO). This work leads to great improvements in finding the best merge opportunity, lateral and longitudinal merge planning and control, trajectory postprocessing and driving comfort. Our algorithm's robustness, adaptiveness and efficiency have been tested and validated both in simulations and on-road tests.

## I. INTRODUCTION

In the past several decades, autonomous driving has made a great process extending the ability of autonomous vehicle from Adaptive Cruise Control (ACC) and Lane Keeping Assistance (LKA) to more complex functions such as automatic lane changing, off-road navigation, stopping at the traffic light, etc. [1][2]. In merge scenarios, however, the vehicle cannot make the best merge choice comparing with humans who can predict other vehicles intentions and react intelligently. To deal with such scenario, prediction and planning should be well coupled considering driving efficiency, safety, and comfort. Therefore, a robust and intelligent algorithm that can interact with human-operated traffic on freeways is still under research and development. There are mainly two approaches in dealing with automatic merge problem:

One category is the prediction and planning coupled method which converts the prediction and planning problem into a convex optimization form, and prediction information is transferred into inequality constraints [3]. Mixed Logical Dynamics (MLD) [4][5] is also applied in solving this optimization problem. [6] uses quintic polynomials in merge planning, while polynomial parameters are chosen as optimization variables. This approach often lacks computational efficiency, and its robustness is not well guaranteed. Another category is the prediction and planning decoupled method which first finds the best merging opportunity, and then plans a merge maneuver. [7] uses Prediction and Cost Function Based (PCB) method to find the best merge opportunity, while its acceleration and velocity outputs are discontinuous. [8] uses a rule-based planner to decide when to merge based on its pre-defined state machine, but it may fail when facing complex traffic scenarios.

In terms of motion planning, there are also two major approaches: path/speed decoupled method and path/speed coupled method. The first approach separately plans velocity and path and combine they together to get spatial trajectory with time information. While the second approach separately plan lateral and longitudinal trajectories and combine them together to get spatial trajectories. In some path/speed decoupled methods, polynomial curvature spiral is generated, and spatial trajectory is solved by using Lagragian method [9][10][11]. Cubic polynomial is also used to generate path [12][13]. In the two methods, piece-wise velocity profile is then generated satisfying some road and path constraints. In merge problem, path/speed decoupled method does not work well since merging maneuver requires that the vehicle reaches a position at a specific timestamp (position/time strictly coupled). In path/speed coupled methods, quintic polynomials are generated longitudinally and laterally, and they are combined to generate spatial trajectories [14][15]. Such method, however, does not consider comfort and merge opportunity when dealing with merge problem. Therefore, it may lead to an uncomfortable lateral acceleration in lane changing, and even leads to danger if an unreasonable merge opportunity is selected.

In trajectory postprocessing, gradient descent method is used in [16] to optimize trajectory curvature, but it fails to do collision checking after trajectory smoothing. A Dual-Loop Iterative Anchoring Path Smoothing (DL-IAPS) is used to smooth trajectory generated by hybrid A* algorithm, by converting the trajectory smoothing problem into a convex optimization problem. The algorithm has an average running time about 0.18-0.21 seconds [17]. However, optimization-based algorithm may degrade system instantaneity in the merge problem, where vehicles usually move at high speed.

In this paper, we propose a novel path/speed coupled method for automatic merge planning. More specifically, we decoupled the method into two hierarchical steps: Dual-Layer Automatic Merge Planning (DL-AMP) and Descent-Based Trajectory Optimization (DBTO). Our method addresses above mentioned issues with the following advantages:

1) **The Best Merging Opportunity**: In our DL-AMP, the first layer is a prediction and cost function-based algorithm which can find the best merging opportunity. The difference between our method and [18] is that our method is only used in finding the best merging opportunity. When the autonomous vehicle begins to merge, the second layer algorithm will be triggered. Therefore, the output continuity is guaranteed by the second layer algorithm. Meanwhile, since the task is much easier for the first layer algorithm, it has less cost function parameters to tune. In our method, candidate ego vehicle acceleration and time are sampled in some ranges. Based on the sampled acceleration and time, constant acceleration and exponential acceleration prediction models are used to predict the relative distance and velocity between ego vehicle and other traffics.

[1] Authors who contribute equally to the article.

All Authors are with Shanghai Automotive Industry Corporation, Ltd. 201, Anyan Road, Jiading, Shanghai, China. *Corresponding author: Yuncheng Jiang   jiangyuncheng@saicmotor.com

Cost function that emphasizes driving efficiency and comfort is designed, and the maneuver that has the minimum cost is chosen satisfying safety and vehicle dynamic constraints.

*2) Efficient Optimization- and Sampling- based Method*: although the quintic polynomial sampling method has been used in [14], it does not well deal with merge scenario motion planning. EM planner [22] generates optimal trajectory by solving iteratively dynamic and quadratic programming problems (DP & QP) with an average of 259*ms*. Unlike EM planner, in our DL-AMP, the second layer algorithm is an optimization- and sampling- based path/speed quintic polynomial trajectory generation method. We only sample on time, and under each fixed time, we convert the trajectory generation problem into a constrained quadratic programming (QP) problem and solve it efficiently. After a suboptimal trajectory is generated under each sampling time, another cost function is formulated that considers sampling time. The optimal trajectory is the one with the lowest cost function. Our optimization- and sampling- based method does not have DP process and piece-wise polynomial trajectory is simplified into a single trajectory, thus greatly reducing computation time.

The longitudinal planning is divided into distance planning and velocity planning to adapt to different merge scenarios. We divide merge scenarios into three categories, and different desired longitudinal distances are selected under different scenarios. To improve efficiency, the constraints of sampling time, desired lateral offset in lateral planning and desired distance and velocity in longitudinal planning are well selected based on road geometry, vehicle dynamic constraints and human driver statistics.

*3) Driving Comfort and Control Feasibility*: To improve driving comfort and control feasibility, we do trajectory smoothing after optimal trajectory is generated. A gradient based method that considers trajectory smoothness and curvature is used to reduce trajectory discontinuity and turning radius which results in reduced lateral acceleration and steering wheel angle in control. Since we use DL-AMP to find the best merge opportunity, free space is ensured at the very beginning, and collision checking is not necessary after smoothing. Unlike [19], we also modify our gradient based method by dealing with exploding and vanishing gradient problems.

This paper is organized as follows: DL-AMP and DBTO are illustrated in Section II and III, respectively. The simulation and on-road test results are shown in Section IV, and conclusion and future work are in Section V.

## II. DUAL-LAYER AUTOMATIC MERGE PLANNING

In this section, we introduce the two parts of DL-AMP: In II-A we introduce how to find the optimal merging opportunity and in II-B we introduce merging trajectory generation.

### A. Finding the Optimal Merging Opportunity

Before the ego vehicle begin to change lane, it should first try to adjust its speed (accelerate, decelerate, wait) to find a suitable free space in the destination lane. Therefore, in the first layer of AMP, the optimal merging opportunity is determined using PCB algorithm that guarantees merge efficiency, comfort, and safety. More specifically, the ego vehicle is supposed to move to a position that has enough longitudinal interval between target vehicles with low relative velocity. For convenience, we divide merging scenarios into three types(A-1). We use constant acceleration prediction model for ego vehicle, and exponential acceleration prediction model for other vehicles(A-2). The desired acceleration of ego vehicle and the time of adjustment are then sampled and evaluated according to a well-designed cost function to bring ego vehicle to the appropriate position and velocity before merging(A-3).

*1) Scenario:* Merge scenarios are divided into three types shown in Fig.1: 1). Ego vehicle merges to the back of vehicle $a$. 2) Ego vehicle merges to the middle of vehicle $a$ and vehicle $b$. 3) Ego vehicle merges to front of vehicle $b$.

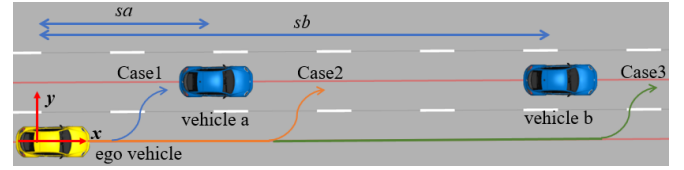

Fig.1: Different merge scenarios. $s_a$ is the distance between ego vehicle and vehicle $a$, and $s_b$ is the distance between ego vehicle and vehicle $b$. $s_{ego}$ is ego vehicle's driving distance.

*2) Prediction:* We use constant acceleration prediction model for ego vehicle and exponential acceleration prediction model for other vehicles.

The distance and velocity of ego vehicle are denoted by

$$s_{ego}(t) = v_0 t + \frac{1}{2} a_0 t^2 \quad (1)$$

$$v_{ego}(t) = v_0 + a_0 t \quad (2)$$

For other vehicles in target lane, we assume that their acceleration would decrease exponentially:

$$\ddot{s}_a(t) = A_0 e^{-\frac{t}{T_a}}, \quad (3)$$

$$\ddot{s}_b(t) = B_0 e^{-\frac{t}{T_b}} \quad (4)$$

$T_a$ and $T_b$ indicate how fast the vehicle acceleration decreases. This hypothesis is derived from the assumption that vehicles in highway are always intended to drive with constant velocity in the long term, even if they have relatively large instant acceleration temporarily.

*3) Cost Functions*: Steady state relative distance cost $C_{dist}$, time to collision cost $C_{ttc}$, total time cost $C_t$, and acceleration cost $C_{acc}$ are chosen to evaluate efficiency, comfort, and safety. We define $P_{dist} = \frac{dist_{ego \to a}}{dist_{a \to b}}$, where $dist_{ego \to a}$ is the distance between ego vehicle and vehicle $a$, and $dist_{a \to b}$ is the distance between vehicle $a$ and vehicle $b$. The steady state position of ego vehicle is supposed to be close to the middle of the other vehicles (in scenarios where there is only one target vehicle, we create another virtual target vehicle). Therefore, $C_{dist}$ increase as $P_{dist}$ deviates from 0.5. $C_{ttc}$ indicates whether it is safe to begin merging by evaluating the time to collision with other vehicles. We assume that time to collision larger than 3*s* guarantees safety. $C_{ttc}$ is given by:

$$C_{ttc} = \frac{1}{2}(C_{ttc}^{rear} + C_{ttc}^{front}) \quad (5)$$

We want to reach the optimal merge point as soon as possible and penalize large total time which is denoted by $C_t$. $C_{acc}$ is the cost of longitudinal acceleration. Maneuvers with large absolute longitudinal acceleration are penalized. The total cost is given by:

$$C_{total} = w_1 C_{dist} + w_2 C_{ttc} + w_3 C_t + w_4 C_{acc} \quad (6)$$

All costs are regularized in [0,1] for sake of parameter tuning, and is shown in Fig.2.

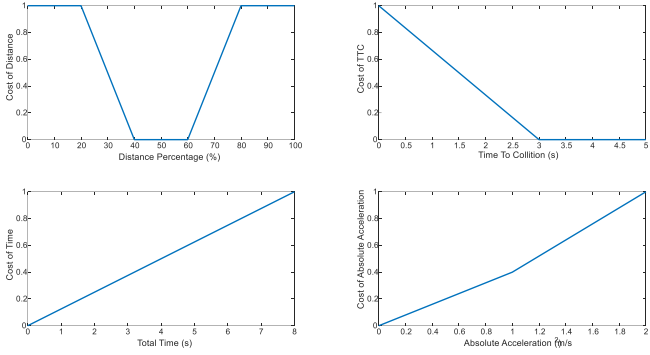

Fig. 2: Cost functions

### B. Merging Trajectory Generation

We use optimization- and sampling- based method to generate optimal quintic polynomials longitudinally and laterally by solving a quadratic programming problem with its standard form:

$$\arg\min_x \frac{1}{2} x^T H x + f^T x$$
$$s.t.\ A_{ieq} x \geq b_{ieq}$$
$$A_{eq} x = b_{eq}$$

We evaluate the candidate trajectories using objective function considering driving efficiency, comfort, and safety. Equality and inequality constraints are applied to guarantee fixed initial and terminal states, vehicle dynamics and driving comfort. To make merge maneuver more like human behaviors, we first determine comfortable lateral acceleration and curvature limits which are shown in Fig.3. It determines the inequality constraints of terminal desired distance in longitudinal distance planning and the constraints of terminal velocity in longitudinal velocity planning.

*1) Lateral Planning:* according to human driving statistics, we limit sampling time in a reasonable range to improve sampling efficiency. Lateral acceleration is the most critical factor that affects driver comfort, and path curvature is also an important factor that affects the performance of lateral controller. Based on the feature of quintic polynomial, we calculate a reasonable time range that satisfies both (Fig.3). The sampling time range is constrained between 4.5s and 7.0s.

In lateral direction, the objective function is denoted by:

$$C_{lat}(l(t)) = K_a \int_0^{T_e} a_t^2(t) dt + K_j \int_0^{T_e} J_t^2(t) dt + K_d \int_0^{T_e} d_t^2(t) dt \quad (7)$$

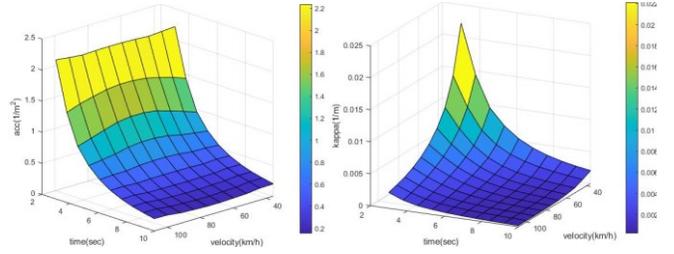

Fig. 3: Lateral features of quintic polynomials. We calculate the maximum lateral acceleration a quintic polynomial can generate with a constant lateral offset (in China, standard road width 3.75m) and longitudinal velocity varying from 36km/h to 108 km/h. We find that the lateral acceleration profile is like a saddle with its peak values (time fixed) ranging between 68km/h and 76km/h. According to driver statistics, we find that lateral acceleration should be less than $1.5 m/s^2$, the minimum time is therefore limited at 4.5s. The curvature graph is then used to check if the maximum curvature is too aggressive with time equal to 4.5s. To ensure merging efficiency, we limit maximum time at 7.0s which is like human drivers' behavior.

where $K_j$, $K_j$, and $K_d$ is penalty weights of different parameters. $a(t)$ is lateral acceleration of the trajectory. $J(t)$ is lateral jerk of the trajectory, and $d(t)$ is lateral offset of the trajectory w.r.t the reference line. Unlike cost function in [6] which contains more than ten parameters, we want to select cost function terms as less as possible, as heavy parameter calibration is always time consuming and does not appeal to engineering application. In terms of comfort, we find that jerk and acceleration are the most direct factors that affects driver comfort. Thus, we only keep acceleration and jerk terms, omitting lateral velocity and acceleration terms. The curvature-related terms are also neglected since we not only have a reasonable sampling time range that ensures proper trajectory curvature, but also do trajectory smoothing in DBTO to optimize curvature. The consecutivity term is neglected in merge scenarios where there exists no symmetry. Less cost function terms can improve computation efficiency and reduce parameter calibration.

We define the initial state constraint $D_0 = \{d_0, \dot{d}_0, \ddot{d}_0, T_0\}$ where $d_0$ is ego vehicle's lateral offset with respect to reference line. $\dot{d}_0$ is ego vehicle's lateral velocity, $\ddot{d}_0$ is ego vehicle's lateral acceleration, and $T_0$ is current time which is always assumed to be zero. The terminal constraints are defined as $D_e = \{d_e, \dot{d}_e, \ddot{d}_e, T_e\}$ where $d_e$ is terminal lateral offset and $T_e$ is terminal time which is fixed in each QP problem. We let $\dot{d}_e = \ddot{d}_e = 0$, as we always want the terminal lateral velocity and lateral acceleration equal to zero. $d_e$ is an inequality constraint which is limited between the boundaries of target road denoted by $RW$. Middle points constraints are performed on these discretized points: $\forall i \in \varphi \mid \varphi = 0, 0.1, 0.2 \ldots T_e]$. The constraints are as follows:

$$\begin{cases} d(0) = d_0, \dot{d}(0) = \dot{d}_0, \ddot{d}(0) = \ddot{d}_0 \\ \dot{d}_{n-1} = 0, \ddot{d}_{n-1} = 0 \\ -\frac{1}{2} RW \leq d(n) \leq \frac{1}{2} RW \\ \dot{d}(i) \in [\dot{d}_{min}, \dot{d}_{max}] \\ \ddot{d}(i) \in [\ddot{d}_{min}, \ddot{d}_{max}] \end{cases} \quad (8)$$

*2) Longitudinal Planning:* longitudinal planning is divided into distance planning and velocity planning. Similar with lateral planning, we use quintic polynomial in distance planning and quantic polynomial in velocity planning. The optimal trajectory is achieved by solving similar constrained QP problem. The objective functions for distance planning and velocity planning are denoted, respectively, by:

$$C_{lond}(s(t)) = K_a \int_0^{T_e} a_t^2(t)dt + K_j \int_0^{T_e} J_t^2(t)dt + K_{ds} \int_0^{T_e} \Delta s^2 dt \quad (9)$$

$$C_{lonv}(s(t)) = K_a \int_0^{T_e} a_t^2(t)dt + K_j \int_0^{T_e} J_t^2(t)dt + K_{dv} \int_0^{T_e} \Delta v^2 dt \quad (10)$$

where $\Delta s = s_{target} - s(t)$ and $\Delta v = v_{set} - v(t)$.

$K_a, K_j, K_{dv}$ and $K_{ds}$ are respective penalty weights of different parameters. $a_t$ is longitudinal acceleration. $J_t$ is longitudinal jerk, $\Delta s$ is the longitudinal offset between desired distance and terminal distance, and $\Delta v$ is the longitudinal velocity offset between set speed and terminal speed.

In distance planning, we define the initial state constraints $S_0 = \{s_0, \dot{s}_0, \ddot{s}_0, T_0\}$ where $s_0, \dot{s}_0$, and $\ddot{s}_0$ are respectively ego vehicle's longitudinal distance with respect for reference line, longitudinal velocity, and longitudinal acceleration. $T_0$ is current time which is always assumed to be zero. The terminal state constraints are defined as $S_e = \{s_e, \dot{s}_e, \ddot{s}_e, T_e\}$ where $s_e, \dot{s}_e, \ddot{s}_e$ are terminal longitudinal distance, velocity, and acceleration, respectively. $T_e$ is the same as that in lateral planning. In velocity planning, the terminal states constraints are a little bit different: $S_e = \{\dot{s}_e, \ddot{s}_e, T_e\} = \{\dot{s}_e, 0, T_e\}$, we leave $s_e$ as unconstrained. Middle points constraints are performed on these discretized points: $\forall i \in \varphi \mid \varphi = 0, 0.1, 0.2 \ldots T_e]$. The desired longitudinal distance $s_{target}$ is calculated differently according to different scenarios (II-A), we use Case I as an example for our next contents and follow the exponential acceleration prediction model.

By solving boundary value problems, the position, velocity and acceleration of both vehicle $a$ and $b$ $(s_a, s_b, \dot{s}_a, \dot{s}_b, \ddot{s}_a, \ddot{s}_b)$ can be easily achieved. We omit the calculation for brevity.

Ego vehicle is supposed to move to the middle of the other two target vehicles, and $s_{target}$ is denoted by:

$$s_{target} = s_a + \frac{1}{2}(s_b - s_a) = \frac{1}{2}s_a + \frac{1}{2}s_b \quad (11)$$

The first order and second order derivatives of $s_{target}$ are as follows:

$$\dot{s}_{target}(t) = \frac{1}{2}\dot{s}_a(t) + \frac{1}{2}\dot{s}_b(t) \quad (12)$$

$$\ddot{s}_{target}(t) = \frac{1}{2}\ddot{s}_a(t_0) + \frac{1}{2}\ddot{s}_b(t_0) \quad (13)$$

In distance planning, the constraints are as follows:

$$\begin{aligned} s(0) = s_0, \dot{s}(0) = \dot{s}_0, \ddot{s}(0) = \ddot{s}_0 \\ \dot{s}_{n-1} = \dot{s}_{target}(T_e), \ddot{s}_{n-1} = \ddot{s}_{target}(T_e) \\ 0.8 s_{target} \leq s(n) \leq 1.2 s_{target} \\ \dot{s}(i) \in [\dot{s}_{min}, \dot{s}_{max}] \\ \ddot{s}(i) \in [\ddot{s}_{min}, \ddot{s}_{max}] \end{aligned} \quad (14)$$

In longitudinal velocity planning, the constraints are as follow:

$$\begin{aligned} s(0) = s_0, \dot{s}(0) = \dot{s}_0, \ddot{s}(0) = \ddot{s}_0 \\ \dot{s}_{n-1} = \dot{s}_{target}(T_e), \ddot{s}_{n-1} = \ddot{s}_{target}(T_e) \\ V_{min} \leq \dot{s}(n) \leq V_{max} \\ \dot{s}(i) \in [\dot{s}_{min}, \dot{s}_{max}] \\ \ddot{s}(i) \in [\ddot{s}_{min}, \ddot{s}_{max}] \end{aligned} \quad (15)$$

The complete formulation of the optimization problem is as follows:

$$\tilde{C}_{total} = \tilde{C}_{lat}(l(i)) + \tilde{C}_{lond}(s(i)) \quad (16)$$

subject to lateral constraints and longitudinal distance constraints or velocity constraints according to different longitudinal scenarios. After optimal trajectory is generated at each sampling time $T$, total cost function is formulated and the trajectory with minimum cost is selected as global optima:

$$C_{total} = \tilde{C}_{total} + K_T T \quad (17)$$

Although we can formulate the total cost function in another form that contains $t$:

$$C_{total}(s(t)) = C_{lat}(l(t)) + C_{lond}(s(t)) + t \quad (18)$$

The first two terms are in quadratic form, while $t$ becomes an optimization variable, the objective function becomes a nonlinear form accordingly, and we can only use gradient-based method or nonconvex optimization methods such as SQP (Sequential Quadratic Programming) to deal with the problem. The computation time greatly increases, and we will lose algorithm efficiency in engineering applications.

On the other hand, we can solve the above problem by extreme search. In real scenario tests, we found that to ensure the solution quality, it is necessary to keep high end state sampling density, which leads to high computation time. Although loose sampling density takes about an average of 0.1ms, which is like the computation time QP method takes, the computation time increases quadratically as sampling density increase, and takes up to 3ms in the worst case.

III. DESCENT BASED TRAJECTORY OPTIMIZATION

In this section, to further reduce the curvature and heading angle of the optimal quintic polynomial, we introduce our descent-based trajectory optimization, and the overall algorithm is shown in **Algorithm 1**. Our method is a post-processing of trajectory generated by DL-AMP. It is based on gradient descent which is an iterative optimization algorithm. The coordinate of each point is updated iteratively in direction of the negative derivative of the objective function. In our method, the objective function is a weighted sum of three terms: curvature, smoothness, and straightness which is given by:

$$C_{smooth} = w_c \sum_{i=1}^{N}(k_i - k_{max})^2$$

$$+ w_{straight}\sum_{i=1}^{N}\Delta x_i^2 + w_{smooth}\sum_{i=1}^{N}(\Delta x_{i+1} - \Delta x_i)^2 \quad (19)$$

$w_c$, $w_{straight}$, and $w_{smooth}$ are penalty weights of different terms. The first term is penalty on curvature. The second term is penalty on straightness, and the third term is penalty on smoothness where $\Delta x_i$ is defined as $x_{i+1} - x_i$. Post-processing of trajectory generated by Hybrid A* is introduced in [20][21], but quintic polynomial has less noise than trajectory generated by Hybrid A* algorithm. Therefore, we have not seen expected gradient descent from the smoothness term in real tests. Accordingly, we introduce another straightness term to deal with the vanishing gradient problem. We also introduced a buffer band that is close to the original quintic polynomial to avoid the bad effect of exploding gradient. To prevent trajectory points' heading angles that are close to initial and terminal locations from moving considerably, variable penalty weights which have Gaussian distribution are introduced. To guarantee the robustness of the trajectory optimization algorithm, we set three stopping criteria: 1). Maximum iteration number; 2). Buffer band; 3) Minimum curvature. The maximum iteration number ensures that the optimization algorithm is forced to stop at fixed maximum allowed running time. The buffer band ensures that the optimized trajectory is near the original trajectory. It can not only check the explode gradient problem, but also keep the validity of collision free status. Minimum curvature criteria stop the algorithm once the trajectory is smooth enough for lateral controller.

### A. Vanishing Gradient Problem

In objective function, we introduce smooth term. In merge scenarios, longitudinal distance change is much longer than lateral distance change, making gradient descent in smooth term very small and making the smoothing algorithm very inefficient. Although we can use bigger descent step, it should adapt to different merge scenarios, and sometimes unreasonable step size may cause explode descent problem. To ensure obvious gradient descent in trajectory smoothing, we introduce another term $\Delta x_i^2$. Human drivers intend to turn steering wheel as slowly and less as possible, which indicating that the merge trajectory becomes smoother and straighter. We use a simple merge scenario to show the effect of straightness term. In this case, DL-AMP generates an optimal trajectory with a maximum lateral offset of 3.5m and a longitudinal offset of 65m. We compare the effect of smoothness term for trajectory optimization with descent step size 0.15 and allowing 400 iterations. The result is shown in Fig.4.

**Algorithm 1**: Gradient Descent Algorithm

**Variable:**
id: trajectory index.
$\nabla$: gradient of trajectory points.
$c_{max}^{traj}$: the maximum absolute value of trajectory curvature
$d_{max}^{traj}$: the maximum absolute value of offset between original trajectory and smoothed trajectory
**Parameters:**
$\alpha, \beta, \gamma$: descent step size of smoothness, straightness, curvature term.
$\sigma$: the standard deviation of Gaussian distribution.
$L$: the length of trajectory points.
$P$: original trajectory points $[P_x, P_y]$.
$B$: buffer band.
$c_{max}$: satisfactory curvature.

```
1    COND ← 1
2    While COND = 1 do
3        for all P∈ original trajectory do
4            ∇ ← (0,0)
5            α ← GaussianDistribution (id, σ, L)
6            ∇ ← ∇ − SmoothnessTerm(x_{i−2}, x_{i−1},
7                  x_i, x_{i+1}, x_{i+2})
8            ∇ ← ∇ − StraightnessTerm(x_i, x_{i+1})
9            ∇ ← ∇ − CurvatureTerm(x_{i−1}, x_i, x_{i+1})
10           x_i ← x_i + ∇
11       end for
12       i ← i + 1
13       if i > MAX_ITER or c_max^traj < c_max or d_max^traj > B then
14           COND ← 2
15           break
22   end while
```

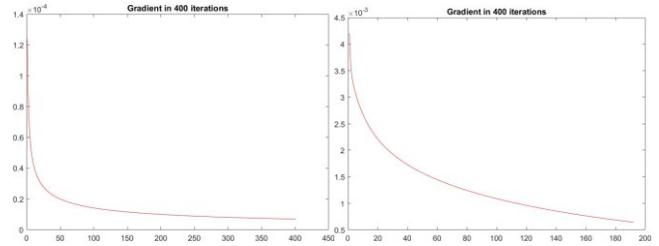

Fig.4 Effects of straightness term. The gradient with $\Delta x_i^2$ term (right figure) is at least twenty times bigger than that without $\Delta x_i^2$ term (left figure). In this case, the iteration is much more efficient and stops at the 191st iteration saving at least half of the computational time.

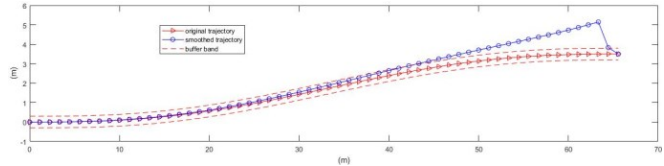

Fig.5: Optimized trajectory without constraints from buffer bands.

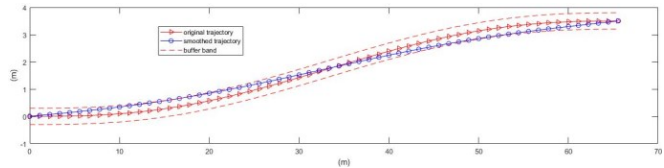

Fig.6: Optimized trajectory with constraints from buffer bands.

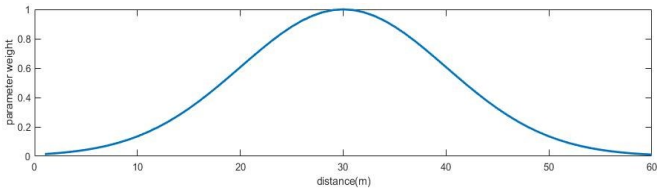

Fig.7: Variable weights with Gaussian distribution

*B. Buffer Band*

Buffer band is introduced to avoid explode gradient problem and keep the validity of collision free status. As is shown in Fig.5. and Fig.6. gradient of smoothness term explodes if a big step size is selected, thus the optimized trajectory becomes unreasonable. We use buffer band to avoid gradient exploding.

*C. Variable Penalty Weights*

To overcome the problem of abrupt difference of heading angle close to the initial and terminal points, we prevent points close to the initial and terminal locations of the trajectory from moving considerably. The gradient descent step size is modified as a variant parameter with Gaussian distribution. This variant parameter controls the step size with respect for the point index. As a result, points near the initial and terminal points have smaller gradient descent step size, thus moving less compared to those points far from the initial and terminal points. The Gaussian distribution is shown in Fig. 7.

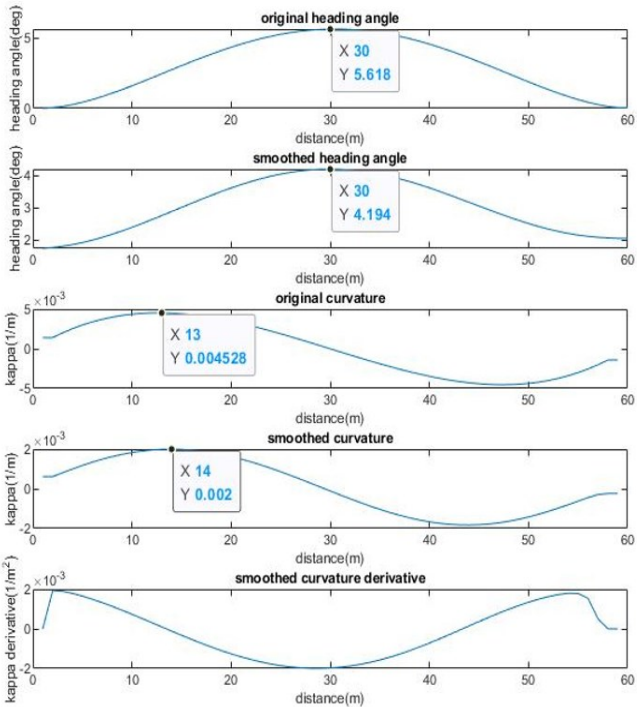

Fig. 8: Trajectory features before and after smoothing. Gradient based smoothing improves the characteristic of trajectory heading angle and curvature. Variable weight of step length with Gaussian distribution ensures the continuity of optimized heading angle close to initial and terminal positions of original trajectory. The continuity of change rate of curvature is still guaranteed after trajectory smoothing.

*D. Gradient Descent*

The gradient of each term is as follows: (The detail of derivation of $k_i$ can be find in [20])

1). **The Derivative of Smoothness Term**

$$\frac{\partial}{\partial x_i}\left[w_s \sum_{i=1}^{N}(\Delta x_{i+1} - \Delta x_i)^2\right] \quad (20)$$
$$= x_{i-2} - 4x_{i-1} + 6x_i - 4x_{i+1} - x_{i+2}$$

**2). The Derivative of Curvature Term**

The partial derivative of $k_i$ with respect to $x_i$ is:

$$k_i = \frac{\Delta \phi_i}{\Delta x_i} \quad (21)$$

where $\Delta \phi_i$ is the angle between $\Delta x_i$ and $\Delta x_{i+1}$:

$$\Delta \phi_i = \cos^{-1}\frac{\Delta x_i^T \Delta x_{i+1}}{|\Delta x_{i+1}||\Delta x_i|} \quad (22)$$

3). **The Derivative of Straightness Term**

$$\frac{\partial}{\partial x_i}\left[w_s \sum_{i=1}^{N}(\Delta x_i)^2\right] = 2x_i - x_{i-1} - x_{i+1} \quad (23)$$

*E. Smoothing Results*

We use the same test parameters as that in section A to see the performance of DBTO. In Fig. 8, the optimized trajectory is well improved in terms of heading angle and curvature. The maximum heading angle is reduced from 5.618 *deg* to 4.194 *deg*, and the maximum curvature is reduced from $0.004528\frac{1}{m}$ to $0.002\frac{1}{m}$. The continuity of heading and curvature are also well reserved due to continuous step size that subjects to Gaussian distribution. The continuity of curvature derivative is also checked to see whether the optimized trajectory is still second order continuous.

## IV. EXPERIMENTAL RESULTS

The proposed motion planning method is tested by Software in Loop (SIL) on Carmaker Platform. Meanwhile, it is also implemented on an IBM-PPC-750GL 900MHz PC with 16M RAM. All road test experiments are carried out on this platform. In this section, we provide two typical and common scenario tests: 1) two vehicles move at speed of 30km/h-40km/h with an interval of 20m-40m. Ego vehicle moves at speed of around 40km/h, and merges into the two vehicles from front. 2). two vehicles move at speed around 40km/h with an interval of 20m-40m. Ego vehicle moves at speed of 30km/h-40km/h, and merges into the two vehicles from behind. Depictions of the road and the trajectories of the vehicles of the two scenarios are shown in Fig.9.

Simulation test results and road test results are shown in Fig. 10 and Fig. 11. We select four variables to indicate the performance of merge maneuver.1) longitudinal interval between ego vehicle and other vehicles at the point that ego vehicle starts merging, indicating whether an optimal merge opportunity is found. 2) steady state longitudinal intervals between ego vehicle and other vehicles, which indicates whether merge maneuver converges asymptotically. 3) ego vehicle longitudinal velocity which indicates longitudinal comfort. 4) ego vehicle lateral velocity which indicated lateral comfort.

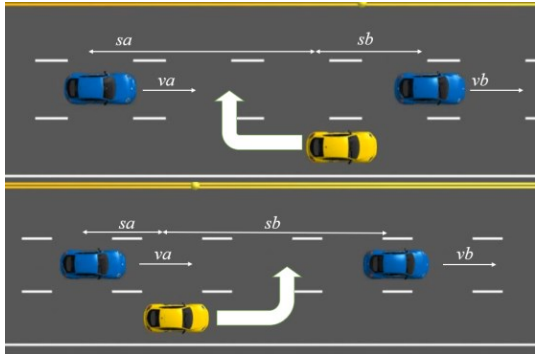

Fig.9: Depictions of the two merge scenarios: in the first scenario, ego vehicle will decelerate, and wait for an open space before merge; in the second scenario, ego vehicle will accelerate to "open up" space before merge.

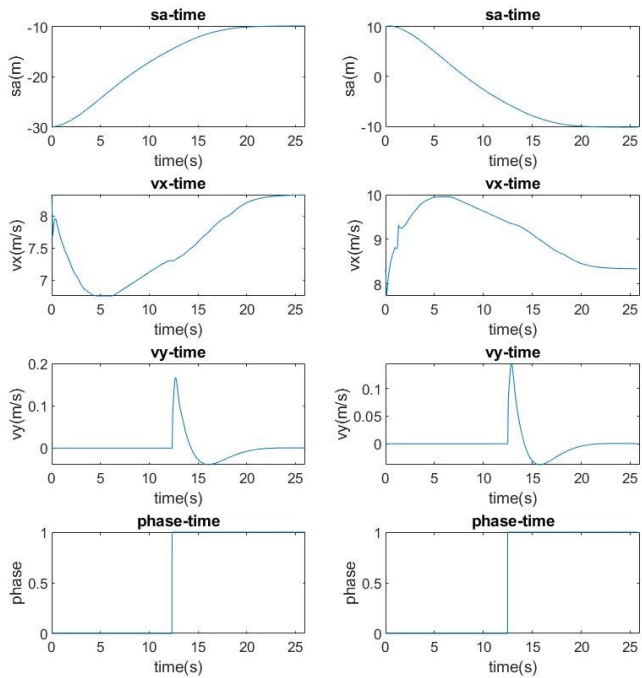

Fig.10: simulation results. The first column is the first scenario test results, and the second column is the second scenario results. In the first scenario, ego vehicle starts at $s_a = -30m$, $s_b = -10m$, $v_{ego} = 30km/h$, $a_{ego} = 0\ m/s^2$. Ego vehicle is expected to slow down and merge into the other two vehicles on the adjacent lane from front. Ego vehicle first decelerate to search for the best merge opportunity. The fourth figure in the first column indicates that the optimal merge opportunity is found when phase number becomes one. At 12.4s, the best opportunity is found, and ego vehicle has longitudinal interval $s_a = -14.6m(s_b = 5.4m)$, $v_{ego} = 26.2km/h$. Then ego vehicle begins to accelerate and merge, and finally moves to the middle of the other two vehicles with $v_{ego} = 30km/h$. Lateral velocity has maximum value of 0.16m/s. In the second scenario, ego vehicle starts at $s_a = 10m$, $s_b = 30m$, $v_{ego} = 30km/h$, $a_{ego} = 0\ m/s^2$. Ego vehicle is expected to speed up and merge into the other two vehicles on the adjacent lane from behind. Ego vehicle first accelerates to search for the best merge opportunity. At 12.5s, the best opportunity is found, and ego vehicle has longitudinal interval $s_a = -5.5m(s_b = 14.5m)$, $v_{ego} = 33.8km/h$. Then ego vehicle begins to decelerate and merge, and finally moves to the middle of the other two vehicles with $v_{ego} = 30km/h$. Lateral velocity has maximum value of 0.14m/s.

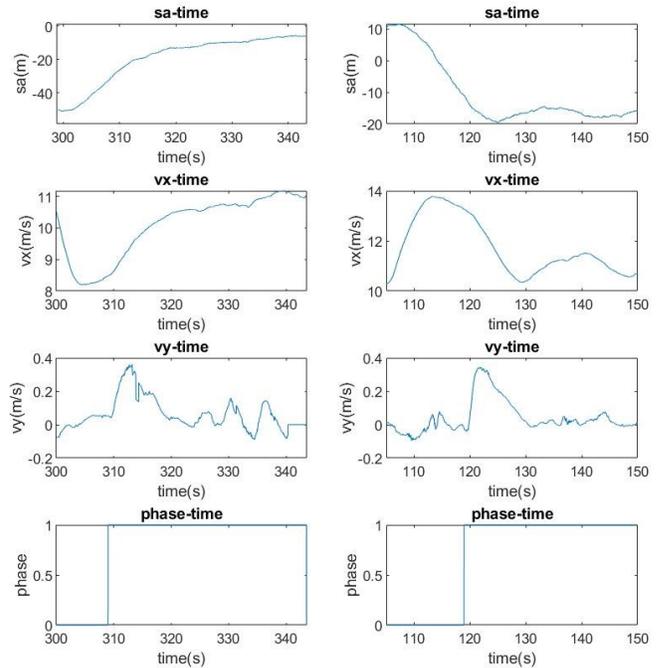

Fig.11: Road test results. The first column is the first scenario test results, and the second column is the second scenario results. In the first scenario, ego vehicle starts at $s_a = -50m$, $s_b = -10m$, $v_{ego} = 36km/h$, $a_{ego} = 0\ m/s^2$. Ego vehicle is expected to slow down and merge into the other two vehicles on the adjacent lane from front. Ego vehicle first decelerates to search for the best merge opportunity. The fourth figure in the first column indicates that the optimal merge opportunity is found when phase number becomes one. At 309 s, the best opportunity is found, and ego vehicle has longitudinal interval $s_a = -27m(s_b = 13m)$, $v_{ego} = 30.24km/h$. Then ego vehicle begins to accelerate and merge, and finally moves to the middle of the other two vehicles with $v_{ego} = 40km/h$. Lateral velocity has maximum value of 0.35m/s. In the second scenario, ego vehicle starts at $s_a = 11.29m$, $s_b = 52m$, $v_{ego} = 40km/h$, $a_{ego} = 0\ m/s^2$. Ego vehicle is expected to speed up and merge into the other two vehicles on the adjacent lane from behind. Ego vehicle first accelerates to search for the best merge opportunity. At 118.9s, the best opportunity is found, and ego vehicle has longitudinal interval $s_a = -9.7m\ (s_b = 30m)$, $v_{ego} = 48.06km/h$. Then ego vehicle begins to decelerate and merge, and finally moves to the middle of the other two vehicles with $v_{ego} = 40km/h$. Lateral velocity has maximum value of 0.33m/s.

## V. CONCLUSION AND FUTURE WORK

We present a novel optimization- and sampling-based automatic merge algorithm. This method is divided automatic merge into two stages: in the first stage, an improved PCB algorithm is used to find the optimal merge opportunity. Very few cost function terms are selected to ensure its computational efficiency; in the second stage, optimal trajectory is efficiently generated by modeling motion planning as an optimization- and sampling-based problem. Constraints are well selected based on road geometry, vehicle dynamic constraints and human driver statistics to ensure computation efficiency. The computation is with average total computation time 0.3ms.

In the future, we plan to use data-driven method to improve the best opportunity finder so that it becomes more intelligent and is more like human decision behavior. We also plan to use apprenticeship learning to find human driver merge patterns and automatically tune the planner parameter so that the planner output can better mimic human drivers.